\documentclass{article}

\PassOptionsToPackage{square,numbers}{natbib}

\usepackage[preprint]{neurips_2022}

\usepackage[utf8]{inputenc} 
\usepackage[T1]{fontenc}    
\usepackage{hyperref}       
\usepackage{url}            
\usepackage{booktabs}       
\usepackage{amsfonts}       
\usepackage{nicefrac}       
\usepackage{microtype}      
\usepackage{xcolor}         

\usepackage{authblk}
\usepackage{graphicx}
\usepackage{adjustbox}
\usepackage{multirow}
\usepackage{hyperref}
\usepackage{booktabs}
\usepackage{todonotes}
\usepackage[normalem]{ulem}

\definecolor{turq}{rgb}{0.286, 0.658, 0.678}
\definecolor{iml}{rgb}{0.678, 0.286, 0.529}
\definecolor{tum}{rgb}{0.094, 0.352, 0.768}
\definecolor{bblue}{rgb}{0.0, 0.58, 0.71}

\usepackage{pifont}
\usepackage{subcaption}

\newcommand{\old}[1]{\unskip}
\newcommand{\new}[1]{#1}

\newcommand{\cmark}{\color{green}\ding{51}}%
\newcommand{\xmark}{\color{red}\ding{55}}%

\title{What do we learn? Debunking the Myth of Unsupervised Outlier Detection}

\author[1, 2, 3]{Cosmin I. Bercea}
\author[1, 4, 5]{Daniel Rueckert}
\author[1, 2, 6]{Julia A. Schnabel}

\affil[1]{Faculty of Informatics, Technical University of Munich, Germany}
\affil[2]{Institute of Machine Learning in Biomedical Imaging, Helmholtz Center Munich, Germany} 
\affil[3]{Helmholtz AI, Helmholtz Munich, Ingolstädter Landstraße 1, D-85764 Neuherberg, Germany}
\affil[4]{Klinikum Rechts der Isar, Munich, Germany}
\affil[5]{Imperial College London, United Kingdom}
\affil[6]{School of Biomedical Engineering and Imaging Sciences, King's College London, United Kingdom}

\begin{document}

\maketitle

\begin{abstract}
Even though auto-encoders (AEs) have the desirable property of learning compact representations without labels and have been widely applied to out-of-distribution (OoD) detection, they are generally still poorly understood and are used incorrectly in detecting outliers where the normal and abnormal distributions are strongly overlapping. In general, the learned manifold is assumed to contain key information that is only important for describing samples within the training distribution, and that the reconstruction of outliers leads to high residual errors. However, recent work suggests that AEs are likely to be even better at reconstructing some types of OoD samples. In this work, we challenge this assumption and investigate what auto-encoders actually learn when they are posed to solve two different tasks. First, we propose two metrics based on the Fréchet inception distance (FID) and confidence scores of a trained classifier to assess whether AEs can learn the training distribution and reliably recognize samples from other domains. Second, we investigate whether AEs are able to synthesize normal images from samples with abnormal regions, on a more challenging lung pathology detection task. We have found that state-of-the-art (SOTA) AEs are either unable to constrain the latent manifold and allow reconstruction of abnormal patterns, or they are failing to accurately restore the inputs from their latent distribution, resulting in blurred or misaligned reconstructions. We propose novel deformable auto-encoders (MorphAEus) to learn perceptually aware global image priors and locally adapt their morphometry based on estimated dense deformation fields. We demonstrate superior performance over unsupervised methods in detecting OoD and pathology. 
\end{abstract}


\section{Introduction}
\begin{figure}
    \centering
    \includegraphics[width=\linewidth]{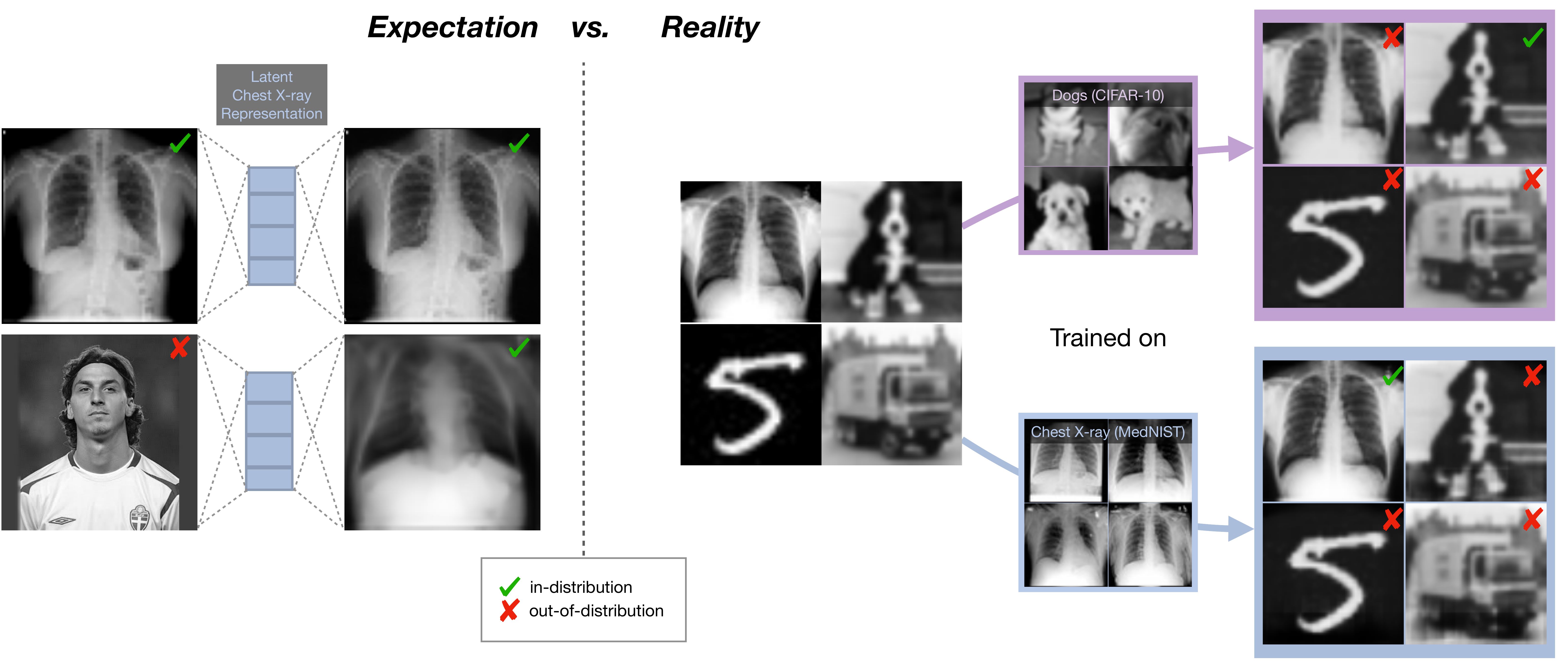}
    \caption{Unsupervised outlier detection assumes that AEs learn the distribution of the training data and therefore only reconstruct samples that match this distribution. For example, an AE trained on chest X-rays would project an OoD sample, e.g., celebrity onto a chest radiography. However, as shown on the right, AEs can reconstruct both in- and out-of distribution samples.}
    \label{fig:intro-ae}
\end{figure}
Outlier detection, also referred to as anomaly- or out-of-distribution (OoD) detection, involves identifying samples or patterns that do not conform to the expected normal distribution. Anomalies are irregular and typically rare data instances, making it difficult to collect large amounts of annotated samples that would cover the full abnormality spectrum. Therefore, supervised~\cite{ciresan2012deep, kamnitsas2017efficient, vyas2018out} and self-supervised~\cite{golan2018deep, kascenas2022denoising} methods can only capture limited facets of the abnormal distribution~\cite{anomaly_review, ruff2021unifying}.

Consequently, the most common methods for anomaly detection are based on unsupervised approaches~\cite{pimentel2014review}, with three popular classes of methods: (1) One-class classifiers~\cite{perera2019ocgan, sabokrou2018adversarially}, where the goal is to determine whether a new sample belongs to the same given class; (2) probabilistic models~\cite{abati2019latent, kingma2018glow, pinaya2021unsupervised, van2016conditional} that use the likelihood scores from the trained models to detect outliers directly; and (3) reconstruction-based methods~\cite{baur2021autoencoders, SCHLEGL201930, somepalli2021unsupervised}, which are optimized to reconstruct only in-distribution samples well, and thus detect anomalies from inaccurate reconstructions of abnormal samples. However, probabilistic and reconstruction-based methods, might assign higher likelihood values, or can better reconstruct OoD samples~\cite{choi2018waic,ImprovingAEMahalanobis, nalisnick2018deep, ren2019likelihood}, with the likelihood being dominated by low-level features, that are common to all natural image data-sets~\cite{schirrmeister2020understanding}. While this can be advantageous for some tasks, such as for reconstruction~\cite{schlemper2017deep}, or restoration~\cite{mao2016image}, anomalies are not detected due to small residual errors. In~\autoref{fig:intro-ae} we show that AEs trained on dogs, or chest X-ray images, are able to reconstruct OoD samples such as trucks, or digits as well. Similarly, Perera et al.~\cite{perera2019ocgan} showed that an AE trained on the digit 8 can also reconstruct digits from the classes 1,5,6 and 9.

In this work, we focus on the latter class and investigate whether state-of-the-art (SOTA) deep reconstruction-based AEs can learn meaningful manifold representations for the task of OoD detection on samples coming from different domains. In addition, we investigate whether they can learn the healthy anatomy (i.e. absence of pathology), generating pseudo-healthy reconstructions of abnormal samples, enabling not just detection but pixel-wise localization of abnormality. Our findings are that SOTA unsupervised methods either do not efficiently constrain the latent space, allowing the reconstruction of anomaly, or the decoder cannot accurately restore images from their latent representation, leading to blurry reconstructions. AEs trained with an adversarial loss can generate images from the learned healthy distribution, but lack pixel-wise reconstruction accuracy, resulting in high residual errors on both healthy and abnormal regions. 

Very recently Zhou et al.~\cite{zhou2022rethinking} investigated the limitations of reconstruction-based AEs for OoD. Similarly, we believe that reconstruction-based AEs should have two properties: i). the latent space should be constrained, so only the reconstruction of in-distribution samples is possible and ii). the decoder should have sufficient capacity to reconstruct in-distribution samples from their latent representations with high accuracy. In contrast to~\cite{zhou2022rethinking}, where the authors aim at reconstructing only low-dimensional feature vectors needed for the classification task, we are interested in reconstructing high-resolution images to enable pseudo-healthy image synthesis and pixel-wise localization of abnormalities. To this end, we propose novel deformable AEs (MorphAEus) to learn perceptually aware global image priors and adapt their shape locally based on estimated dense deformation fields. 

Our manuscript advances the understanding of unsupervised anomaly detection methods by providing insights and quantifying what AEs learn to solve OoD tasks. In summary, our contributions are: 
\begin{itemize}
    \item We highlight the limitations and broaden the understanding of reconstruction-based AEs for unsupervised outlier detection
    \item We investigate whether SOTA AEs can learn the train data distribution, recover inputs from their latent representation with high accuracy\new{,} and reliably detect outliers\footnote{The code will be made public upon acceptance}.
    \item We propose novel deformable AEs (MorphAEus). A constrained AE that learns the training distribution and uses perceptual loss and local morphometric adaptation driven by estimated deformation fields to increase reconstruction accuracy.
\end{itemize}
\section{Unsupervised Outlier Detection. A Myth?}
%
%
The widely held popular belief is that AEs can learn the distribution of the training data and identify outliers from inaccurate reconstructions of abnormal samples~\cite{ruff2021unifying}. Do AEs actually learn the training distribution or is it a myth? 
In this section, we highlight the common underlying assumptions and present the most popular neural networks architectures used for OoD. 
%
%
\subsection{Manifold Learning: Assumptions}
AEs aim to extract meaningful representations from data, by learning to compress inputs to a lower-dimensional manifold and reconstruct them with minimal error.
Let $\mathcal{X} \subset \mathbb{R}^N$ be the data space that describes normal instances for a given task. The manifold assumption implies that there exists a low-dimensional manifold $\mathcal{M} \subset \mathbb{R}^D \subset \mathcal{X}$ where all the points $x\in\mathcal{X}$ lie, with $D\ll N$. For example, a set of images in pixel space $\mathcal{X}$ could have a compact representation describing features like structure, shape, or orientation in $\mathcal{M}$.

Given a set of unlabeled data $x_1, .., x_n \in \mathcal{X}$ the objective of unsupervised manifold learning is to find a function $f:\mathbb{R}^N\rightarrow\mathbb{R}^D$ and its inverse $g:\mathbb{R}^D\rightarrow\mathbb{R}^N$, such that $x\approx g(f(x))$, with the mapping $f$ defining the low-dimensional manifold $\mathcal{M}$. The core assumption of unsupervised anomaly detection is that once such functions $f$ and $g$ are found, the learned manifold $\mathcal{M}$ would best best describe the normal data samples in $\mathcal{X}$ and produce high reconstruction errors for data-points $\overline{x} \notin \mathcal{X}$, that we call anomalous points or outliers. An anomaly score is therefore usually derived directly from the pixel-wise difference between an input and its reconstruction: $s(x) = |x-g(f(x))|$. 

The normal and abnormal distributions are considerably separated from each other when $\overline{x}$ is from a different domain. However, for relevant computer vision tasks, such as industrial inspection, anomalies are often defects in otherwise normal images. In medical imaging, the set $\mathcal{X}$ usually describes the normal/healthy anatomy and the data set $\overline{\mathcal{X}}$ usually contains images with both healthy and pathological regions. The two distributions usually come from the same domain and depending on the abnormalities type and size might overlap considerably. The core assumption is that only the normal/healthy structures can be reconstructed from their latent representation very well, with the pathological regions ideally replaced by healthy structures. Therefore $x~\approx g(f(\overline{x})) \in \mathcal{X}$ would represent the healthy synthesis of the abnormal sample $\overline{x}$ and the residual $|\overline{x}-g(f(\overline{x}))$ would highlight just the abnormal regions.
%
%
\subsection{Auto-Encoders}
Auto-Encoders have emerged as a very popular framework for unsupervised anomaly detection~\cite{gong2019memorizing, pawlowski2018unsupervised, zong2018deep}. They use neural networks to learn the functions $f$ and $g$, often denoted as encoder $E_{\theta} $ with parameters $\theta$ and decoder $D_\phi$ parameterized by a set of parameters $\phi$. The embedding $z=E(x|\theta)$ is a projection of the input to a lower-dimensional manifold $\mathcal{Z}$ also referred to as the bottleneck or latent representation of $x$. The standard objective of AE is finding the set of parameters $\theta$ and $\phi$ that minimize the residual: $\min_{\theta,\phi}\sum_{i=1}^N \| x_i - D_\phi(E_\theta(x_i))\|^2$, with the mean squared error being a popular choice for the reconstruction error. Apart from the reduced dimensionality of the bottleneck, several other techniques are used to regularize auto-encoders~\cite{makhzani2013k, rifai2011contractive}. De-noising auto-encoders (DAEs)~\cite{kascenas2022denoising, vincent2010stacked, zimmerer2019context} explicitly feed noise-corrupted inputs $\tilde{x} = x + \epsilon$ to the network with the aim at reconstructing the original input $x$. This is especially beneficial when the noise model is known a-priori. Variational Auto-Encoders (VAEs)~\cite{burgess2018understanding, Higgins2017bvae, kingma2013auto, zimmerer2019unsupervised} estimate the distribution over the latent space that is regularized to be similar to a prior distribution, usually a standard isotropic Gaussian. Generative adversarial neural networks (GANs)~\cite{goodfellow2014generative} have also been applied to anomaly detection~\cite{perera2019ocgan, SCHLEGL201930}. They learn to fool a discriminator network that the generated images are indistinguishable from real samples, achieving incredible results in high-resolution image generation. Pidhorskyi et al.~\cite{pidhorskyi2018generative} trained auto-encoders with an adversarial loss to detect out-of-distribution samples. More recently, introspective variational auto-encoders~\cite{daniel2021soft, huang2018introvae} use the VAE encoder to differentiate between real and reconstructed samples, achieving outstanding image generations.

\section{MorphAEus: Deformable Auto-Encoders}
\begin{figure}
    \centering
    \includegraphics[width=\linewidth]{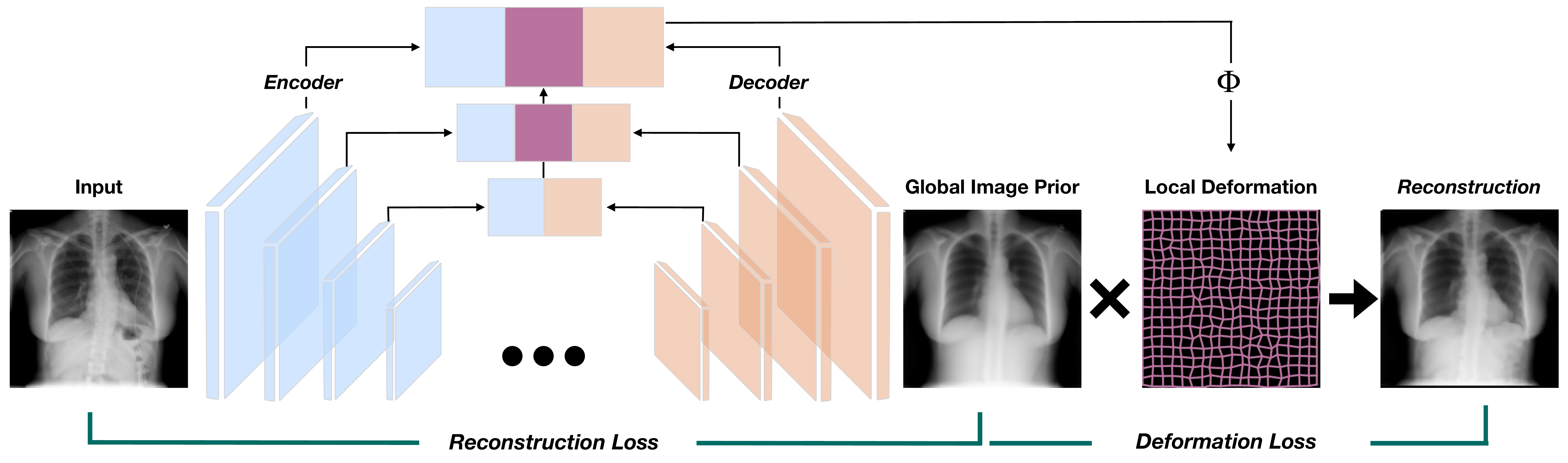}
    \caption{\textbf{MorphAEus: architecture overview.} We leverage deep AEs to predict global image priors matching the training distribution and estimate deep deformation fields from shared encoder-decoder parameters to locally adapt the  morphometry of the prior to match the input.}
    \label{fig::ref-ae}
\end{figure}
In the introduction, we presented the desired properties of any AE for unsupervised anomaly detection, namely i). reconstructions of any input should match the training distribution and ii). the decoder has sufficient capacity to accurately restore the input.~\autoref{fig::ref-ae} illustrates our proposed method. 

\textbf{Global Prior.} AEs use a reduced bottleneck to learn compressed features of an input space. Besides the lower-dimensional bottleneck, other regularization techniques based on sparsity~\cite{vincent2010stacked}, specific noise models~\cite{zimmerer2019context, kascenas2022denoising} or adversarial training~\cite{daniel2021soft, pidhorskyi2018generative} are used. Here, we propose to use deep auto-encoders constrained to reconstruct only in-distribution samples, see~\autoref{sec::intuition} for a visual intuition. Given a dataset $\mathcal{X}=\{x_1, .., x_n\}$ we optimize the encoder and decoder with parameters $\theta, \phi$ to minimize a distance loss, here we use the mean squared error (MSE) between the input and its reconstruction. In addition to the pixel-wise loss, we optimize a perceptual loss (PL)~\cite{johnson2016perceptual} to encourage perceptually similar reconstructions. The reconstruction loss is given by: 
\begin{equation}
    \mathcal{L}_{rec}(x|\theta;\phi) = MSE(x, x_{prior}) + \alpha PL\textrm{, with }x_{prior} = D_{\phi}(E_{\theta}(x)),
\end{equation}
where $\alpha$ weights the perceptual term. We find a value of $\alpha=0.05$ to be a good value to predict perceptually similar images, without compromising pixel-wise accuracy.

\textbf{Local Deformation.} Similar to using skip connections~\cite{ronneberger2015u}, we propose to make use of the upper layers to recover spatial information. However, the direct use of skip connections would allow the network to bypass the learned prior from the bottleneck and copy anomalies at inference time. Instead, we propose to use the shared features from encoder and decoder to locally match the reconstruction to the input. Inspired by the advances in image registration~\cite{balakrishnan2019voxelmorph} and computer vision~\cite{chen2021unsupervised, shu2018deforming} we estimate dense deformation fields $\Phi$ to allow local morphometric adaptations of the estimated prior. The deformation objective is given by:
\begin{equation}
\mathcal{L}_{warp}(x|\psi;\theta_S;\phi_S) = LNCC(x_{warp}, x) + \beta \|\Phi\|^2  \textrm{, with }x_{warp} = x_{prior} \circ \Phi,
\end{equation}
where $\theta_S, \phi_S$ are the shared parameters of the encoder and decoder, LNCC is the local normalized cross correlation, $\circ$ is a spatial transformer and $\beta$ weights the smoothness constraint on the deformation fields. Deformable registration between normal and pathological samples is in itself an active area of research. While beneficial on healthy images, the deformation could potentially mask structural and photometric abnormalities at inference if not constrained. In our experiments, we linearly increase $\beta$ from $1e^{-3}$ to $3$ to more strongly constrain the deformation as the estimation of the global prior improves. The full objective is given by the two losses: $\mathcal{L}(x|\theta;\phi;\psi) = \mathcal{L}_{Rec}(x|\theta; \phi) + \mathcal{L}_{warp}(x|\psi;\theta_S;\phi_S)$. We train the networks jointly, but introduce the deformation loss after 10 epochs ensure a good initializer for the deformation estimation.

\textbf{Network architecture}
We used simple convolutional AEs with a kernel size of $3\times 3$, batch normalization and swish activations. For the inputs of size $128\times128$ in~\autoref{sec::pathology}, the encoder has 7 layers with (16, 32, 64, 128, 256, 256, 256) filters with max-pool operations for down-sampling, projecting the inputs to a latent representation of size $1\times1\times128$. The decoder is analogous to the encoder with bilinear up-sampling between the layers to increase the spatial resolution. The deformation estimation network consists of three transposed convolution layers with $32$ filters and a kernel size of $3\times 3$, batch normalization, and swish activation.
\section{Do We Learn the Manifold? On Out of Distribution Detection\label{sec::ood}}
\begin{table}
    \caption{We report the structural similarity (SSIM) and perceptual difference (LPIPS) to assess the reconstruction accuracy; the average FID score ($\overline{FID}$), and the average classifier confidence ($\overline{Conf}$) to quantify the networks capability of only reconstructing in-distribution samples; and the AUROC to assess the outlier detection performance. Visual examples are shown in~\autoref{fig:benchmark_manifold}.\label{tab::benchmark_sota}}
    \centering
    \setlength{\tabcolsep}{4pt}
     \begin{minipage}{\textwidth}
        \begin{adjustbox}{width=\textwidth,center} 
            \begin{tabular}{c | l | c c | c c | c c c c c c }
                \toprule	    
        	    & \multirow{3}{*}{Method} & \multicolumn{2}{c|}{Reconstruction} &  \multicolumn{2}{c|}{Manifold} & \multicolumn{6}{c}{OoD Detection}  \\
        	    & & \multicolumn{2}{c|}{Accuracy} & \multicolumn{2}{c|}{Learning} & CCT & AbdomenCT & HeadCT & Hand & BreastMRI & MNIST\\
                & & SSIM $\uparrow$ & LPIPS $\downarrow$ & $\overline{FID} \downarrow$ & $\overline{Conf} \uparrow$ & \multicolumn{6}{c}{AUROC $\uparrow$} \\\hline
        	    \xmark & Spatial AE~\cite{baur2021autoencoders} & \bfseries 0.956 & \bfseries0.012 & 283 & 0.15 & 0.920 & \bfseries0.994 & \bfseries0.994 &\bfseries 0.996 & \bfseries0.999 & \bfseries1.0\\
        	    \xmark & Dense AE~\cite{baur2021autoencoders} &0.829 & 0.304 & 255 & 0.49 & 0.936 & \bfseries0.995 & 0.982 & \bfseries0.997 &\bfseries 0.994 &\bfseries 0.995\\
        	    \xmark & VAE~\cite{kingma2013auto, zimmerer2019unsupervised} & 0.830 & 0.289 & 312 & 0.31 & 0.140 & 0.613  & 0.956 &\bfseries 0.994 & 0.983 & 0.984\\
        	    \xmark & $\beta$-VAE~\cite{burgess2018understanding, Higgins2017bvae}& 0.654& 0.350 & 287 & 0.41 & \bfseries0.998 &\bfseries 0.998 & \bfseries0.995 & 0.984 &\bfseries 0.997 &\bfseries 0.995\\
        	    \cmark & AAE~\cite{pidhorskyi2018generative} & 0.725 & 0.112 & 265 & \bfseries0.78 & 0.762 & 0.913  &\bfseries 0.998 &\bfseries 0.993 & \bfseries0.998 &\bfseries 0.998\\
         	    \cmark & S-Intro VAE~\cite{daniel2021soft} & 0.739 & 0.110 & 239 & \bfseries0.74 & 0.852 & 0.952 & 0.987 &\bfseries 0.994 & 0.962 & 0.964 \\\hline
                 \xmark & DAE~\cite{kascenas2022denoising} &0.940 & 0.021 & 277 & 0.07 & 1.0 & 1.0  & 0.998 & 0.974 & 0.995 & 0.907\\\hline
                \cmark & MorphAEus (ours) & 0.850 & 0.099 &  \bfseries{194} & \bfseries{0.76} & \bfseries{0.999} & \bfseries{0.997} & \bfseries{0.998} & \bfseries{0.996} & \bfseries{0.993} & \bfseries{0.996}\\\hline

         	    \bottomrule
            \end{tabular}
        \end{adjustbox}
    \end{minipage}
\end{table}
\begin{figure}
    \centering
    \includegraphics[width=\linewidth]{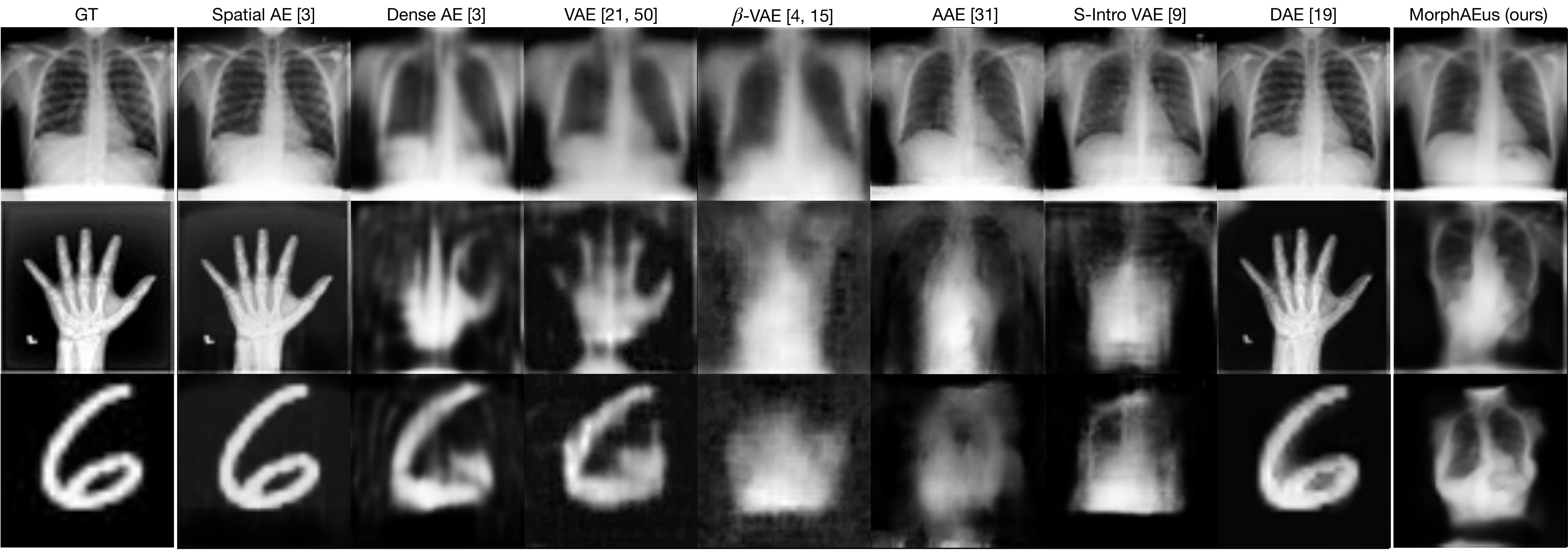}
    \caption{We show reconstructions of in-distribution chest X-rays (top row) and OoD samples (bottom rows). Of the baselines, only AAEs learn the training distribution. Our method produces reconstructions that visually match the training distribution, see.~\autoref{tab::benchmark_sota} for quantitative results.}
    \label{fig:benchmark_manifold}
\end{figure}

In this section\new{,} we investigate whether SOTA auto-encoders learn meaningful manifold representations for the task of OoD detection on samples coming from different domains. 

\textbf{Dataset and metrics.} We used the publicly available medical MNIST (MedNIST) dataset~\cite{medmnistv1}. The dataset contains 58954 medical images belonging to 6 classes – ChestCT(10000), BreastMRI(8954), CXR(10000), Hand(10000), HeadCT(10000), AbdomenCT(10000) at a resolution of $64\times64$ pixels. We used 80\% of the chest X-rays for training, 10\% for validation and the remaining 10\% as in-distribution samples for the evaluation. We treated 1000 random samples from each of the remaining classes and also 1000 random samples from the MNIST~\cite{lecun1998mnist} test dataset as out-of-distribution samples. We calculate the structural similarity (SSIM) and perceptual difference (LPIPS) to asses the reconstruction fidelity of the networks and compare the outlier detection performance using the Area Under the Curve of Receiver Operating Characteristics (AUROC) curve. To quantify whether the networks reconstruct only in-distribution samples for any arbitrary input, we measure the similarity between the reconstructed OoD samples and 1000 random samples from the chest X-ray training set, by computing the average Fréchet Inception Distance (FID)~\cite{heusel2017gans} score. We also show the confidence at which a classifier trained on MedNIST (100\% accuracy), recognizes the reconstructions of OoD samples as in-distribution chest X-rays. We highlight with a green check-mark the methods that achieve reconstructions that are by at least 5\% more similar to CXR than the input distribution and where the classifier is more confident in seeing CXR images than the other classes. 

\textbf{Baselines.} ~\autoref{tab::benchmark_sota} compares the performance of our method with the following baselines: Spatial and dense AEs~\cite{baur2021autoencoders}; variational AEs (VAE) first introduced in~\cite{kingma2013auto} and applied to medical anomaly detection by Zimmerer et al.~\cite{zimmerer2019unsupervised}; $\beta$-VAE~\cite{burgess2018understanding, Higgins2017bvae} that learn a factorization of the latent space; and adversarial AEs (AAE) that use a discriminator to differentiate between real and fake samples~\cite{pidhorskyi2018generative} or are trained in an introspective manner by using only the encoder and decoder~\cite{daniel2021soft}. We also compare our method to recent self-supervised denoising AEs (DAE), that were successfully applied to brain pathology segmentation~\cite{kascenas2022denoising}. However, DAEs are not attempting to detect anomalies based on deviations from normality, but rather remove a trained noise model from the input. Since conceptually they are more akin to supervised methods, we don't consider DAEs as an unsupervised baseline. 

\textbf{Results.} Adversarial AEs are the baseline methods that pass the manifold learning test. This is expected since their training objective is to fool a discriminator that samples generated from noise resemble the training distribution. However, their reconstructions are imperfect, leading to high pixel-wise differences with SSIM scores of just $0.725$ and $0.739$. The remaining methods can be categorized into two classes: methods that copy the input, and methods that cannot decode the latent information to produce good reconstructions. Interestingly, spatial AEs are able to reconstruct the structure of OoD samples very well, as can be seen from the SSIM, LPIPS and visual examples in~\autoref{fig:benchmark_manifold}. However, despite being able to reproduce common low-level features as shown in~\cite{schirrmeister2020understanding}, the networks achieve very good detection scores. This might be due to different intensity ranges of the distributions, or background statistics~\cite{krusinga2019understanding, ren2019likelihood}. Dense AE, VAE and $\beta$-VAE do not have sufficient decoding capacity and produce blurred reconstructions. Interestingly, $\beta$-VAE which achieve the lowest SSIM score of 0.654 achieves the best OoD detection performance among the baselines. 

We find that OoD detection results are generally high, suggesting that there is sufficient difference to the distribution of the other domains. In some cases, e.g., CCT or abdominal CT, where the images are generally smoother, the residual errors are also lower, which is more challenging for some of the baseline methods. Our proposed method passes the manifold learning test and achieves the best FID and classification confidence scores. Unlike adversarial networks, our network is able to reconstruct the inputs more accurately, with an  SSIM score of $0.850$ compared to $0.739$, and consistently achieves near perfect detection results on all datasets. 

\section{Do We Learn the Healthy Manifold? On Pathology Detection\label{sec::pathology}}
\begin{table}[tb]
    \caption{We report the structural similarity (SSIM) and the perceptual difference (LPIPS) to assess the reconstruction accuracy on healthy samples, and the FPR95, FPR99, AUROC, and AUPRC to assess the disease detection performance. Refer to~\autoref{fig:benchmark_anomaly} for qualitative interpretations. \label{tab::benchmark_sota_anomaly}}
    \centering
    \setlength{\tabcolsep}{4pt}
     \begin{minipage}{\textwidth}
        \begin{adjustbox}{width=\textwidth,center} 
            \begin{tabular}{l | c c | c c c c| c c c c}
                \toprule	    
        	    \multirow{3}{*}{Method} &  \multicolumn{2}{c|}{Rec.} & \multicolumn{4}{c}{COVID-19} &\multicolumn{4}{c}{Lung Opacity} \\
        	    & \multicolumn{2}{c|}{Healthy} & \multicolumn{4}{c|}{}&   \multicolumn{4}{c}{} \\
                & SSIM  $\uparrow$ & LPIPS $\downarrow$ & FPR95$\downarrow$ & FPR99$\downarrow$ & AUPRC$\uparrow$ & AUROC$\uparrow$ & FPR95$\downarrow$ & FPR99$\downarrow$ & AUPRC$\uparrow$ & AUROC$\uparrow$ \\\hline
    
        	    spatial AE~\cite{baur2021autoencoders} & \bfseries{0.949} & \bfseries{0.015} & 0.949 & 0.987 & 0.840 & 0.559 & 0.945 & 0.990 & 0.884 & 0.562 \\
        	    dense AE~\cite{baur2021autoencoders}  & 0.735 & 0.409 & 0.934 & 0.972 & 0.818 & 0.550 & 0.959 & 0.984 & 0.845 & 0.446 \\
        	    VAE~\cite{kingma2013auto, zimmerer2019unsupervised} & {0.752} & 0.327 & 0.952 & 0.985 & 0.771 & 0.448 & 0.974 & 0.994 & 0.827 & 0.397 \\
        	    $\beta$-VAE~\cite{burgess2018understanding, Higgins2017bvae} & 0.613 & 0.466 & 0.789 & 0.944 & 0.920 & 0.774 & 0.850 & 0.961 & 0.924 & 0.696 \\
        	    AAE~\cite{pidhorskyi2018generative}  & 0.652 & 0.156 & 0.861 & 0.967 & 0.882 & 0.685 & 0.910 & 0.983 & 0.909 & 0.649\\
         	    S-Intro VAE~\cite{daniel2021soft} & 0.688 & 0.165 & 0.937 & 0.974 & 0.856 & 0.610 & 0.957 & 0.985 & 0.881 & 0.532\\\hline
          	    DAE~\cite{kascenas2022denoising} & {0.965} & 0.011 & 0.228 & 0.446 & 0.984 & 0.951 & 0.579 & 0.879 & 0.958 & 0.826 \\\hline
               MorphAEus (ours) & 0.839 & 0.104 & \bfseries0.578 & \bfseries0.828 & \bfseries0.956 & \bfseries0.868 & \bfseries0.587 &\bfseries 0.843 &\bfseries 0.963 & \bfseries0.839 \\
    	        - w/o $\mathcal{L}_{warp}$ & 0.711 & 0.089 & 0.889 & 0.957 & 0.908 & 0.734 & 0.918 & 0.968 & 0.918 & 0.660\\
         	    - w/o $\mathcal{L}_{warp}$; w/o PL & 0.780 & 0.249 & 0.897 & 0.951 & 0.884 & 0.681 & 0.937 & 0.972 & 0.900 & 0.598\\
         	    \bottomrule
            \end{tabular}
        \end{adjustbox}
    \end{minipage}
\end{table}
\begin{figure}[tb]
    \centering
    \begin{minipage}{\textwidth}
        \includegraphics[width=\linewidth]{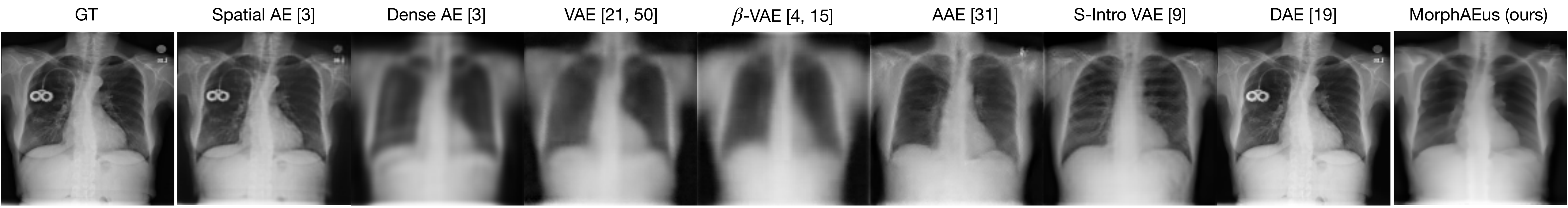}
        \subcaption{Healthy. The spatial and denoising AEs copy the input and are able to reconstruct implanted medical devices as well. Dense AE, VAE, and $\beta$-VAE produce blurry reconstructions. \old{Whereas}\new{While} adversarial methods reconstruct samples similar the healthy training distribution, they lack accurate pixel-wise reconstructions. Our method synthesizes an accurate healthy image, removing the \new{implanted} medical device.}
        \label{fig::healthy}
    \end{minipage}
    \begin{minipage}{\textwidth}
        \includegraphics[width=\linewidth]{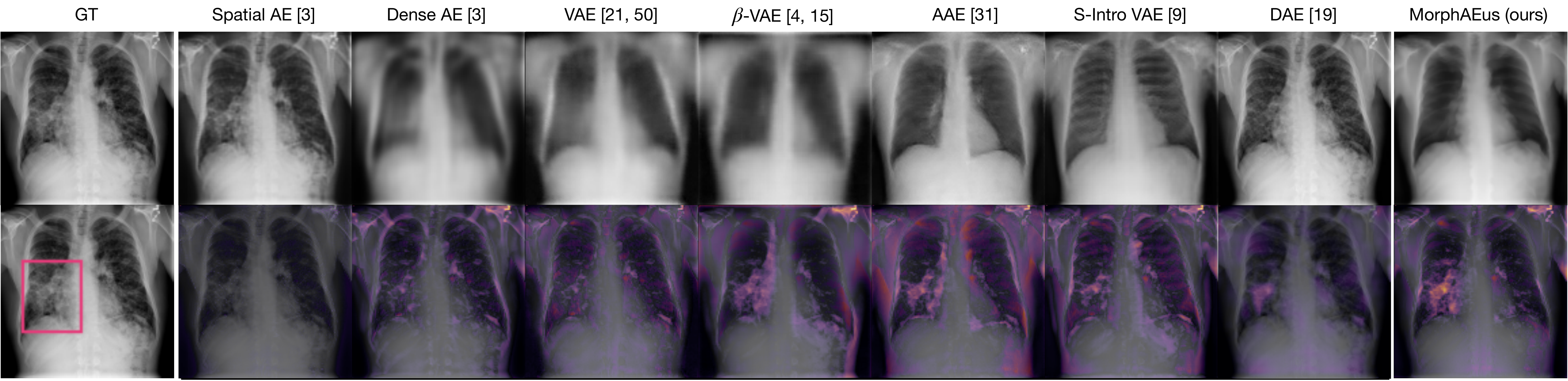}
        \subcaption{Pathology. The first row shows reconstructions and the second row highlights the abnormal regions with overlaid residual heat-maps. Spatial AEs can reconstruct abnormal regions, leading to no residual error. Dense AE, VAE, and $\beta$-VAE also show high residual erros on healthy regions due to the lack of reconstructed details, e.g., ribs. Adversarial AEs detect the anomaly, but also produce high residual due to inaccurate reconstructions. DAEs can remove some of the noise and produce high residual errors on he pathology region. Our approach synthesizes a pseudo-healthy image of the input and is able to accurately detect and localize the abnormality.}
        \label{fig::pathology_normal}
    \end{minipage}
    \begin{minipage}{\textwidth}
        \includegraphics[width=\linewidth]{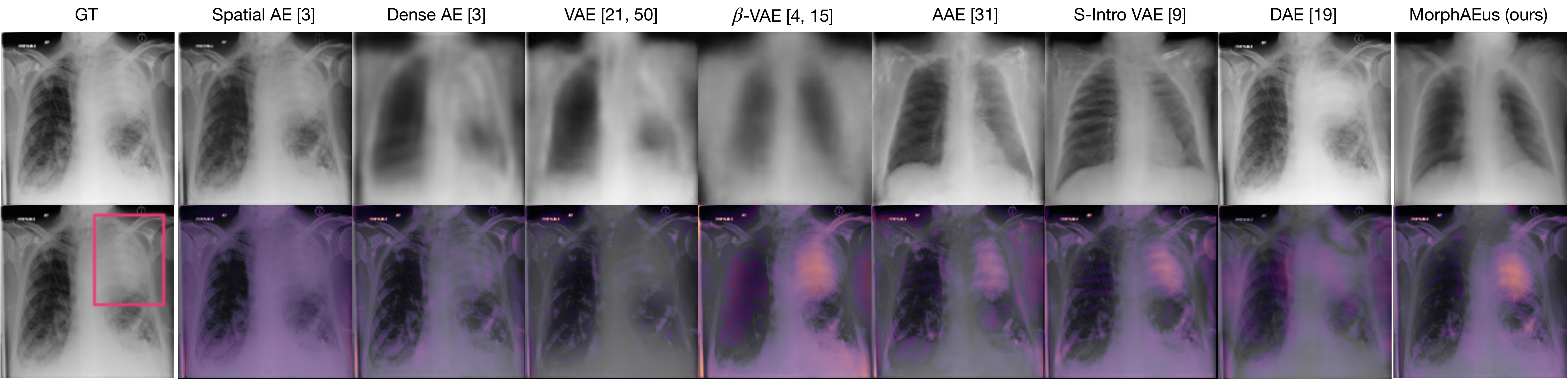}
        \subcaption{Pathology. Severe lung opacity. With almost the entire left lung severely affected by disease, this specific abnormality is not captured by the noise model of DAEs, so they fail to detect it. $\beta$-VAEs, adversarial AEs, and our method induce learned healthy prior and hallucinate lung structure that is not visible. In comparison to the other methods, our method achieves lower residual errors on healthy regions, with the residual heat-map focused around the region affected by the disease.}
        \label{fig::pathology_severe}
    \end{minipage}
    \caption{We show visual examples of healthy samples in~\autoref{fig::healthy}, and pathological cases in~\autoref{fig::pathology_normal} and~\autoref{fig::pathology_severe} with the later one being a more severe case. Refer to~\autoref{tab::benchmark_sota_anomaly} for quantitative evaluation of reconstruction accuracy on healthy samples and detection performance on pathology.}
    \label{fig:benchmark_anomaly}
\end{figure}
In this section\new{,} we investigate whether AEs can learn the healthy anatomy (i.e., absence of pathology) and generate pseudo-healthy reconstructions of abnormal samples to detect COVID-19, and lung opacity pathology on chest X-ray images.

\textbf{Dataset and metrics} We used the publicly available Covid-19 dataset with  normal/healthy(N=10192), Covid-19 (N=3616) and lung opacity (N=6012) chest X-ray images~\cite{dataset_1, dataset_2}. We randomly split the healthy in 80\% training (N=8154), 10\% validation (N=1019) and 10\% test (N=1019) and down-sample all images to $128\times128$ resolution. In addition to the AUROC, we also evaluate \new{ the} following metrics for anomaly detection: Area under the Precision Recall Curve (AUPRC), \new{ and} FPR95/FPR99 being the false positive rates when the true positive rates are at 95\% and 99\% respectively.

\textbf{Results.} ~\autoref{tab::benchmark_sota_anomaly} compares the performance of our method to the baselines presented in~\autoref{sec::ood}. Out of the baseline methods, only adversarial trained auto-encoders can synthesize healthy images from abnormal samples. The other methods either reconstruct abnormal samples, or produce blurry reconstructions, not matching the training distribution.~\autoref{fig::healthy} illustrates this supposition. Spatial AEs and DAEs copy information from the input and are able to reconstruct the medical device, while the next three methods (Dense, VAE, and $\beta$-VAE) produce blurry results, as can be best seen in the LPIPS score.
Overall the outlier detection scores are lower, which is expected since both normal and abnormal samples come from the same domain, and their distributions highly overlap due to common healthy structures.~\autoref{fig::pathology_normal} shows a qualitative example of a pathological sample. As expected, spatial AE can copy the anomaly, which results in almost no residual error. While the next three methods (Dense, VAE, and $\beta$-VAE) have a high residual on the abnormal region, they also produce large residual errors on healthy regions. This is mostly due to the lack of reconstructed details, e.g., on ribs. Even though adversarial networks are able to synthesize normal images from the data distribution, they also produce high residual error on healthy regions, attributed mostly to the variance in the reconstructed shapes. Since the pathology in~\autoref{fig::pathology_normal} lies within the trained noise model of DAEs, they are able to remove some of it in the reconstruction. However, the lung severely affected by disease in~\autoref{fig::pathology_severe} is not captured by the learned Gaussian noise and remains undetected. We show superior performance on detecting COVID-19 and lung opacity compared to unsupervised baselines by a large margin and achieve similar results as the self-supervised approach (DAE) can be seen in~\autoref{tab::benchmark_sota_anomaly}. The the main difference compared to supervised, or self-supervised methods is that we model the healthy anatomy and can detect anomalies beyond the learned labels or noise patterns.

\section{On the Intuition behind MorphAEus\label{sec::intuition}}
We have presented two main ideas that could improve any AE for unsupervised anomaly detection, namely that \old{the reconstructions} \new{AEs} should only produce samples within the training distribution and should also \old{produce} \new{yield} low residual \new{errors} on normal samples. In~\autoref{sec::cor} we show that AEs cannot achieve both and motivate the importance of each of our components. In~\autoref{sec::tails} we discuss the limitations of our method and show samples from the tails of the residual-error distribution (best and worst cases). 
\subsection{Curse of Resolution\label{sec::cor}}
\begin{figure}
    \centering
    \includegraphics[width=\linewidth]{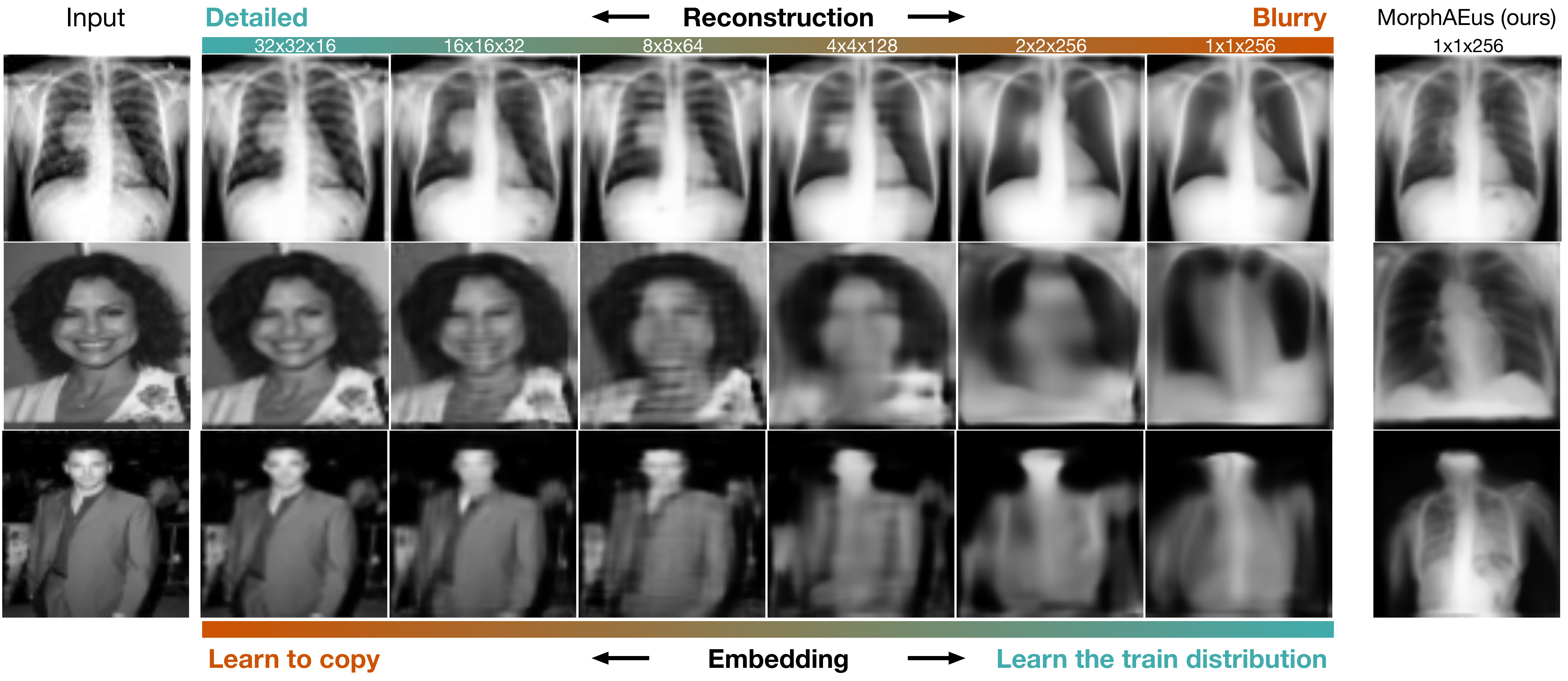}
    \caption{\new{Curse of resolution.} Auto-encoders learn the distribution of the trained dataset (chest X-ray) only at very deep latent levels. However, this comes at the cost of blurry reconstructions that miss details needed for abnormality detection. Our method learns the training distribution and produces detailed reconstructions for both in- and out-of-distribution samples. }
    \label{fig::cor}
\end{figure}
~\autoref{fig::cor} shows reconstructions of samples within the same domain as the training distribution (chest X-rays) but with pathology, and images of celebrities~\cite{liu2015faceattributes} from far out-of-distribution domains. At very shallow levels, \old{auto-encoders} \new{AEs} are able to accurately reconstruct the input, but they do not learn the prior over the training distribution and can thus reconstruct abnormal regions and even far out-of-distribution samples. This leads to poor anomaly detection results\new{,} as it has also been seen for the spatial AE~\cite{baur2021autoencoders} in our experiments in~\autoref{sec::pathology}. Interestingly, with increasing depth, \old{auto-encoders} \new{AEs} can learn the prior over the training distribution, avoid reconstruction of pathology\new{,} and also reconstruct\old{s} the closest chest X-ray counterparts of the celebrities. However, this comes at the cost of losing spatial information and not reconstructing small details, e.g., ribs. This yields high residual for healthy samples, similar to dense AE~\cite{baur2021autoencoders}, VAE~\cite{zimmerer2019unsupervised}, and $\beta$-VAE~\cite{burgess2018understanding} in~\autoref{sec::pathology}\old{, making the networks unusable for anomaly detection}. Concluding, AEs suffer either from copying abnormal patterns\new{,} or \new{from} l\old{o}osing spatial information and details in the reconstruction, making them unsuitable for outlier detection tasks where detailed reconstructions are needed to differentiate between possibly overlapping distribution, e.g., healthy and pathology. Our proposed method produces accurate reconstructions without \old{reconstructions of pathology} \new{restoring pathological structures}, by learning a global image prior that is perceptually similar to the training distribution and locally adapting \old{the} \new{its} morphometry to better match the input. We show an ablation study in the lower part of~\autoref{tab::benchmark_sota_anomaly}, where we remove one component at a time and show the effect on performance. As expected, removing the local deformation module reduces the AUROC scores the most, from 0.868 to \old{0.754} \new{0.734} for COVID-19 detection, and 0.839 to \old{0.659} \new{0.660} for detecting lung opacity. By also removing the perceptual loss, the networks are similar to the deep auto-encoders as in the second-last column in~\autoref{fig::cor} and their detection performance deteriorate further. 
\subsection{Tails of Distribution\label{sec::tails}}
\begin{figure}
    \begin{minipage}{0.48\textwidth}
        \raggedleft
        \includegraphics[width=\linewidth]{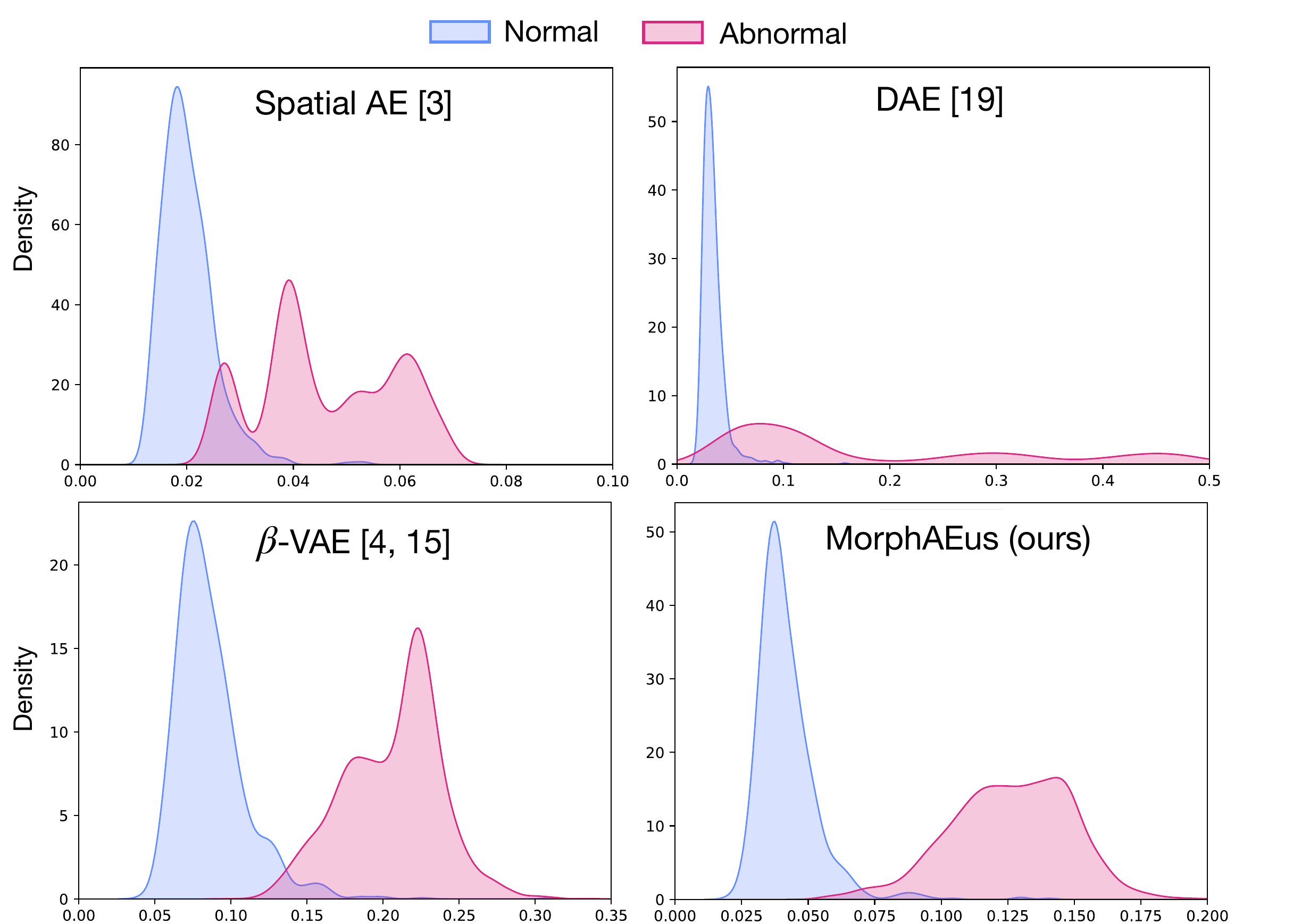}
        \subcaption{\old{OoD experiment in~\autoref{sec::ood}.\label{fig::tails_ood}} \new{OoD performance on MedNIST and MNIST, see~\autoref{sec::ood}. Our proposed method achieves the best separation between normal and abnormal classes.}\label{fig::tails_ood}}
    \end{minipage}
    \begin{minipage}{0.48\textwidth}
    \raggedright
        \includegraphics[width=\linewidth]{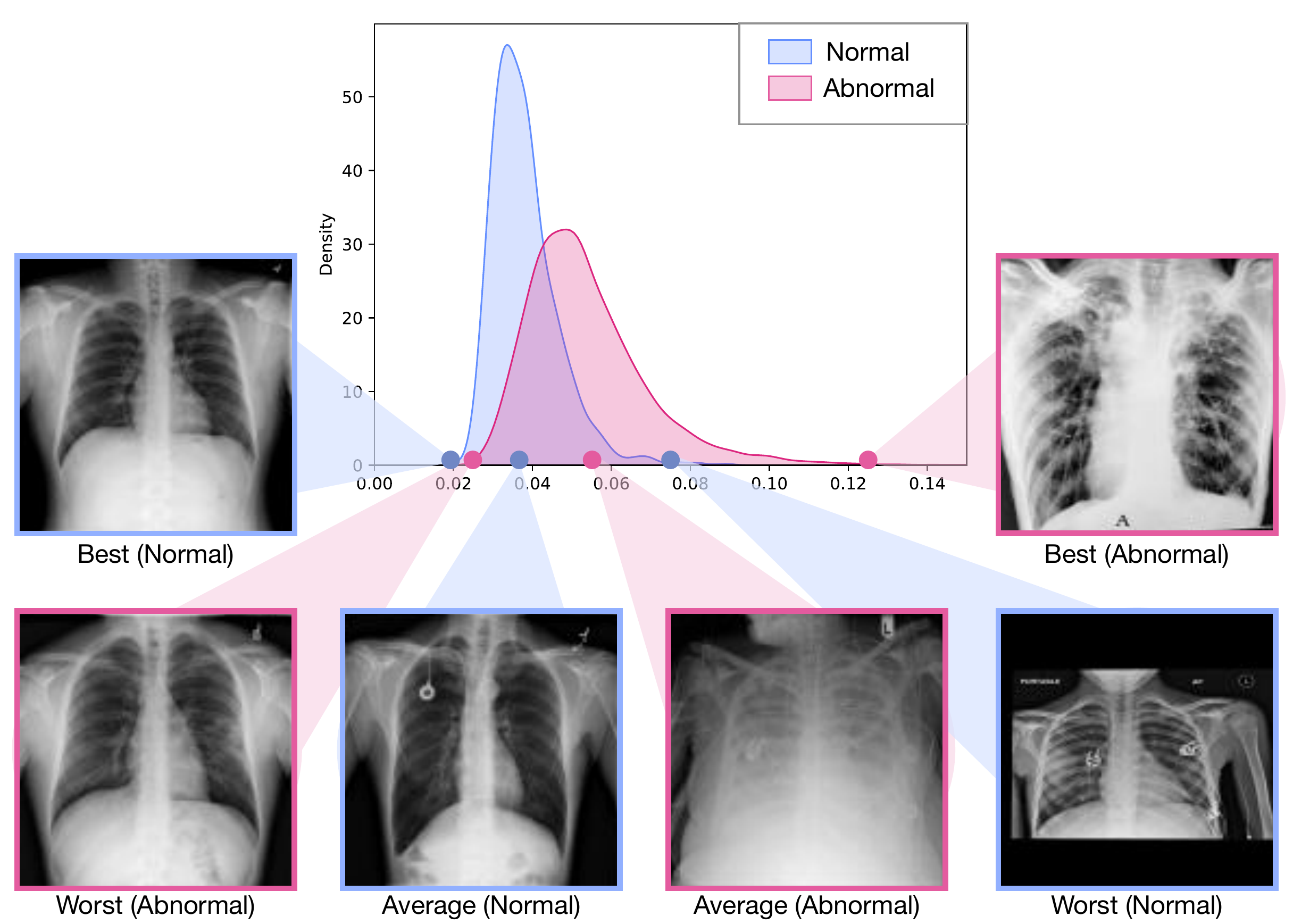}
        \subcaption{\old{Pathology detection analysis, see~\autoref{sec::pathology}.} \new{Pathology detection performance, see~\autoref{sec::pathology}. Our method does not reconstruct artefacts such as text, or implanted devices on images tagged as normal.}}
        \label{fig::tails}
    \end{minipage}
    \caption{Distribution of residual errors of normal and abnormal samples.~\autoref{fig::tails_ood} shows results of the best performing 4 methods in~\autoref{sec::ood}, \new{and~\autoref{fig::tails}} shows best, average and worst results for normal and abnormal distributions of our method \old{for} \new{on} the pathology detection task in~\autoref{sec::pathology}.\label{fig::tails_of_distribution}}
\end{figure}

Here we analyze the performance and limitations of our method.~\autoref{fig::tails_ood} shows density distribution plots of the residual errors for out-of-distribution detection on MedNIST and MNIST in~\autoref{sec::ood}. We compare our method to the best three performing baselines and show that we achieve considerably less overlap between normal and OoD classes \old{with} \new{and} high separation a margin.~\autoref{fig::tails} shows the density plot on detecting pathology, see~\autoref{sec::pathology}. To gain insights in the network failures, we sample the best, average and mean points of the normal and abnormal distributions. Looking at the distribution of the normal cases, our networks produce higher residuals on images tagged as normal, but contain\new{ing} artefacts such as implanted medical devices, text, doctors annotations, or different backgrounds. This has no effect on the clinical relevance \old{of our method}, but plays an important role for the quantitative evaluation, since our method does not reconstruct all of these varying factors and \old{tags} \new{detects} them as anomalous, as can be seen in~\autoref{fig::healthy}. Even more so, the residual errors on text, or medical devices is often much higher than for small pathology. \old{One can curate the test set to include only normal samples without artefacts, or have a classification head to detect text, or implanted medical devices to} \new{To have an accurate impression of the networks performance, the test set can be curated to include only normal samples without artefacts, or an additional classification head to detect text, or implanted medical devices can be employed.} On the same note, we believe that the improved scores of the self-supervised method in~\autoref{sec::pathology} \old{is} \new{are} mostly because it is not affected by this kind of artefacts, as seen in~\autoref{fig::healthy}\new{, and because of high residuals due to failed reconstructions, see~\autoref{fig::pathology_95},~\autoref{fig::pathology_2035},~\autoref{fig::pathology_2025}, and~\autoref{fig::pathology_40}}. Looking at the distribution of residual errors on anomalous samples, we find that the degree of pathology increases with higher residuals, which is the expected behaviour\old{s}. The left tail of the abnormal distribution, shows cases where we achieved the lowest residual scores. We find that \old{this} \new{these} cases contain very small pathology, that are nearly impossible to detect on the small input resolution of $128\times 128$. \new{Processing inputs at a higher resolution is needed} \old{T} \new{t}o also improve the detection of such cases\old{, a solution might be to work on higher resolutions}.
\section{Conclusion}
In this work, we have investigated the use of reconstruction-based AEs for unsupervised outlier detection. We stipulate that any unsupervised anomaly detection method should have the desired property of learning the normal distribution and producing highly accurate reconstructions of samples within the distribution. We have shown that standard, variational, and recent adversarial auto-encoders do not satisfy both conditions and are not suitable for detection tasks where the normal and abnormal distribution highly overlap, e.g., for finding defects in \new{otherwise} normal images or pathological regions in otherwise healthy patients. 
Based on these findings, we propose novel deformable auto-encoders (MorphAEus) that learn a perceptually-aware global image prior and adapt their morphometry to match the input using estimated local dense deformation fields. Our method outperforms SOTA auto-encoders in detecting samples from distant distributions, i.e., different domains, but also in detecting abnormal regions on otherwise healthy samples. Beyond outlier detection, our method could be applied for synthesizing healthy images from pathological samples that can be used as augmentation, or for creating and curating healthy training datasets.

\bibliography{ref}

\newpage
\appendix

\section{Appendix}

\subsection{SOTA implementations}

\textbf{Spatial and Dense AE~\cite{baur2021autoencoders}.} We followed the official public implementation\footnote{https://github.com/StefanDenn3r/Unsupervised\_Anomaly\_Detection\_Brain\_MRI}. We kept the network the same as in the original submission for the CXR anomaly experiments. We adapted the networks to the small input resolution of MedNIST by removing the first block ($64\times64\times32$) and the last block ($128\times128\times32$). As in the original submission we trained the network with a batch size of $128$ using an ADAM optimizer with a learning rate of $1e^{-4}$ and $L1$ reconstruction loss, for up to 1000 epochs with early stopping if no improvement greater than $10e^{-9}$ is achieved for 5 consecutive epochs. We achieved better results with a batch size of $64$, learning rate of $5e^{-4}$ and early stopping for $25$ rounds of no improvement and show these throughout the manuscript.  \\

\textbf{VAE~\cite{zimmerer2019unsupervised}.} We followed the official public implementation\footnote{https://github.com/MIC-DKFZ/vae-anomaly-experiments/}. As in the original implementation we trained the networks with a batch size of $64$, a latent dimensionality of $512$ using an ADAM optimizer with a learning rate of $1e^{-4}$ with a plateau learning rate scheduler (multiplies with 0.1 on plateaus), early stopping after 3 epochs and $L2$ reconstruction loss for up to 1000 epochs. We achieved best results with a learning rate of $5e^{-4}$ and early stopping after $25$ epochs. We adapt the networks to the higher resolution ($128\times128$) of the CXR anomaly experiments by duplicating the last layer with 256 filters. 

\textbf{$\beta$-VAE~\cite{Higgins2017bvae, burgess2018understanding}.} We followed the public implementation\footnote{https://github.com/1Konny/Beta-VAE/}. As in the improved version~\cite{burgess2018understanding}, we set $\beta$ to $4$, $\gamma$ to $1000$ and linearly increased $C$ from $0$ to $50$. The optimizer used was ADAM with a learning rate of $5e^{-4}$. We achieved better results by setting $\gamma$ to 10. For the MedNIST experiments, We used 4 layers with 32 filters as proposed and added one layer with 32 filters to adapt the networks to the higher resolution on the CXR anomaly experiments.

\textbf{DAE~\cite{kascenas2022denoising}.} We followed the official public implementation\footnote{https://github.com/AntanasKascenas/DenoisingAE/}. We adapted the networks to the small input resolution of MedNIST by removing the first block ($64\times64\times64$) and the last block ($128\times128\times64$). As in the original submission we trained the network with a batch size of $16$ using an ADAM optimizer with a learning rate of $1e^{-4}$ and $L1$ reconstruction loss. We trained for up to 1000 epochs and early stopping after $25$ consecutive epochs with no improvement greater than $10e^{-9}$.

\textbf{AAE~\cite{pidhorskyi2018generative}.} We followed the official public implementation\footnote{from https://github.com/podgorskiy/GPND}. We adapted the networks for higher resolutions resolutions by doubling or tripling the first layer of the encoder and discriminator and the last layer of the discriminator for inputs of size $64\times64$ and $128\times128$, respectively. We achieved best results while setting $\mathcal{L}_{error}$ and $\mathcal{L}_{adv-dz}$ when optimizing for $D_z$ to 2.0 and trained for up to 1000 epochs with batch size of $128$, learning rate of $5.0e^{-4}$ and early stopping after $25$ consecutive epochs with no improvement greater than $10e^{-9}$.

\textbf{Soft-Intro VAE (SI-VAE)~\cite{daniel2021soft}.} We followed the official public implementation\footnote{https://taldatech.github.io/soft-intro-vae-web/}. We used 4 layers (64,128,256,512) and 5 layers (64,128,256,512,512) with a bottleneck of 256 filters and a final sigmoid activation for the MedNIST and CXR anomaly experiments, respectively. We set $\beta_{KL}=1, \beta_{Rec}=1, \beta_{Neg}=256$ and used ADAM optimizer with learning rates of $2e^{-4}$ and batch size of $32$ as originally proposed. We trained for up to 1000 epochs and early stopping after $25$ consecutive epochs with no improvement greater than $10e^{-9}$.
\subsection{Pathology Detection: More Visual Examples}
\begin{figure}
    \begin{minipage}{\textwidth}
        \includegraphics[width=\linewidth]{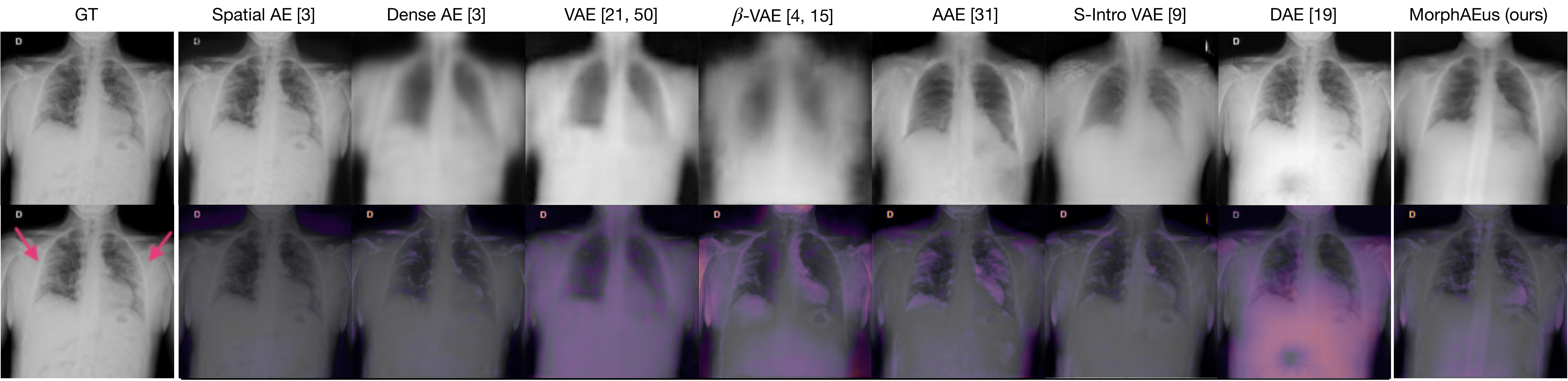}
        \subcaption{COVID-19. $\beta$-VAE and DAE do not accurately reconstruct the input intensity, leading to high image-wise residuals, but no clinical value. }
        \label{fig::pathology_95}
    \end{minipage}
 \begin{minipage}{\textwidth}
        \includegraphics[width=\linewidth]{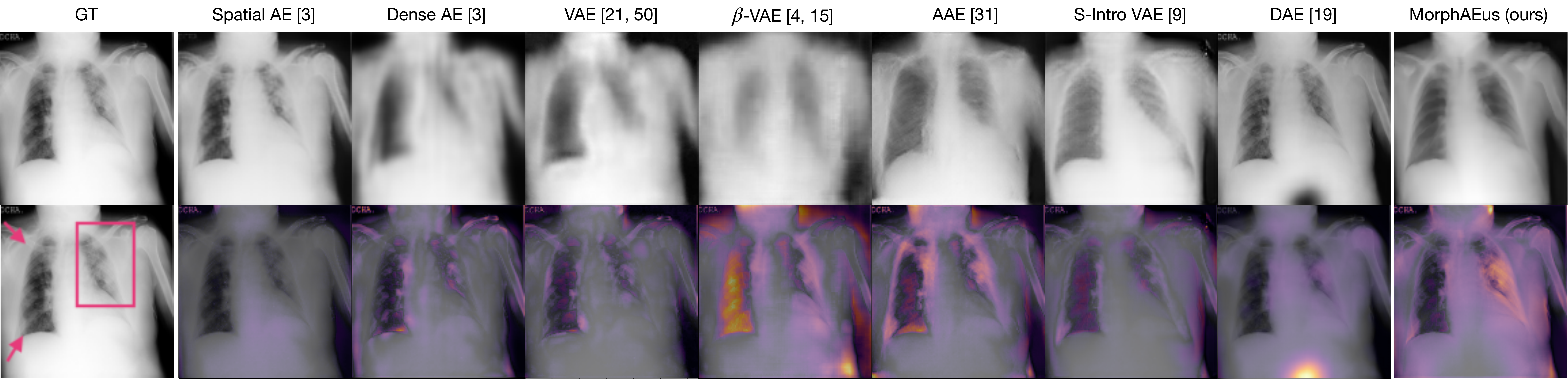}
        \subcaption{COVID-19. DAE achieves high-residual due to generated artefact in the abdomen.}
        \label{fig::pathology_2035}
\end{minipage}
 \begin{minipage}{\textwidth}
        \includegraphics[width=\linewidth]{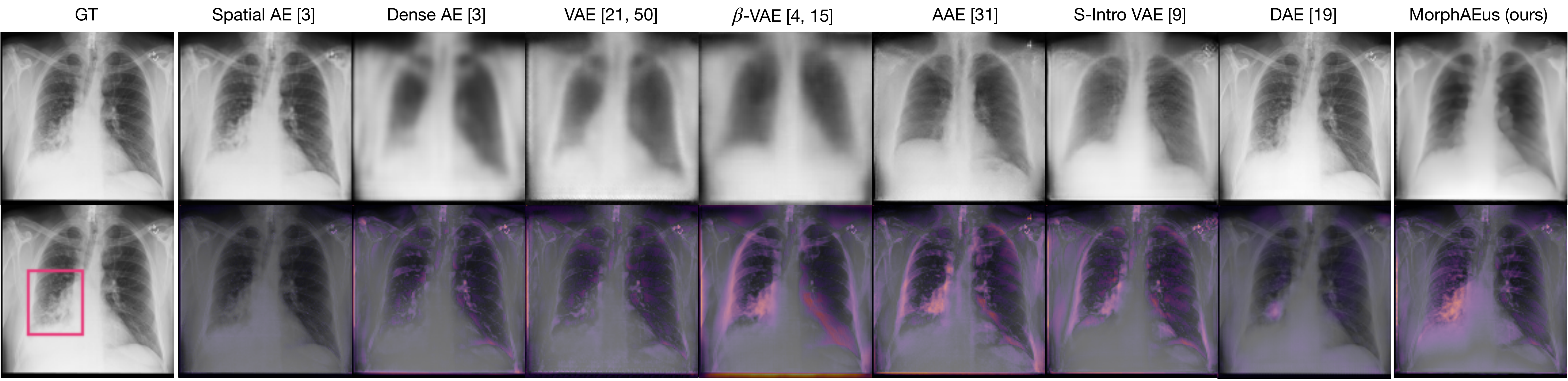}
        \subcaption{Lung opacity. Spatial, Dense, VAE and DAE do not detect the disease. $\beta$-VAE, AAE, and S-Intro VAE achieve more false positives than our proposed method. }
        \label{fig::pathology_2010}
\end{minipage}
 \begin{minipage}{\textwidth}
        \includegraphics[width=\linewidth]{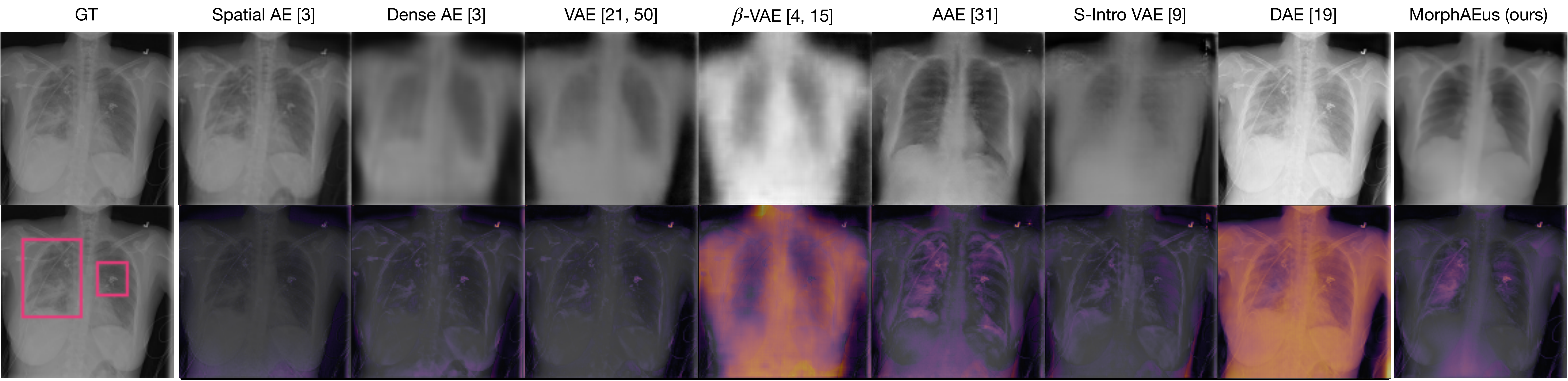}
        \subcaption{Lung opacity. Spatial, Dense and VAE do not detect the pathology. $\beta$-VAE and DAE do not capture the intensity profile, leading to high residuals, but no clinical value. MorphAEus detects the pathology and produces less false positives.}
        \label{fig::pathology_2025}
\end{minipage}
 \begin{minipage}{\textwidth}
        \includegraphics[width=\linewidth]{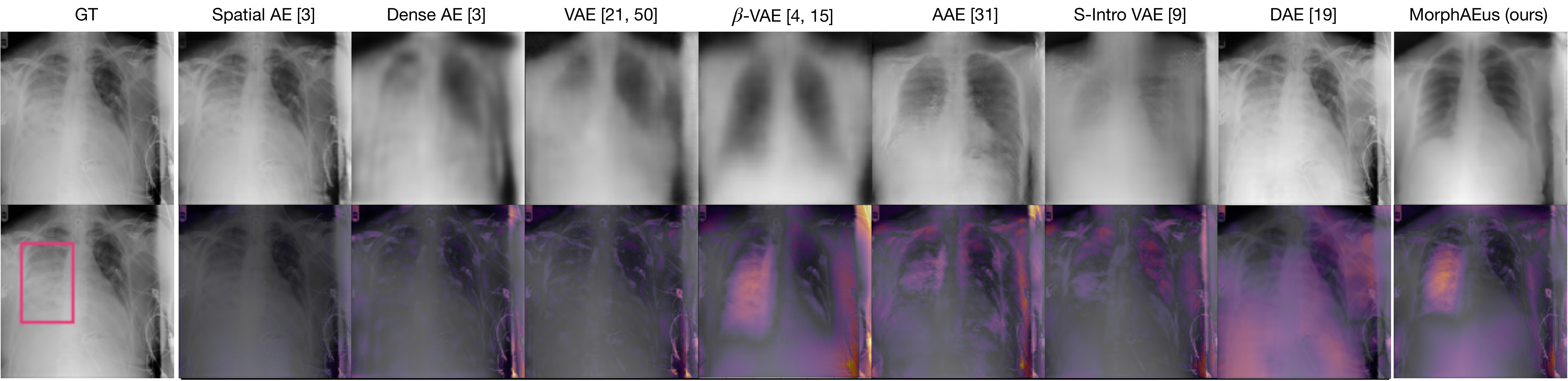}
        \subcaption{Lung opacity. Spatial, Dense and VAE do not detect pathological structures. DAE denoises the input in various areas where there is no pathology present. MorphAEus reconstructs an accurate pseudo-healthy version of the abnormal input, detects the pathology and achieves less false positives.}
        \label{fig::pathology_40}
\end{minipage}
\end{figure}

~\autoref{fig::pathology_95} and~\autoref{fig::pathology_2035} show COVID-19 cases. ~\autoref{fig::pathology_2010},~\autoref{fig::pathology_2025} show mild cases of lung opacity, while~\autoref{fig::pathology_40} and shows a case where the lung has been severely affected by disease. 

\end{document}